\documentclass[10pt,journal,comsoc]{IEEEtran}
\usepackage[utf8]{inputenc}
\usepackage[T1]{fontenc}
\usepackage[normalem]{ulem}
\usepackage[pdftex]{color}
\usepackage{graphicx}% Include figure files
\usepackage{cite}
\usepackage{hyperref}% add hypertext capabilities
\usepackage{mathtools}
\usepackage{amsmath}
\usepackage[capitalise]{cleveref}
\usepackage{booktabs}

\begin{document}

\title{Decentralised Federated Learning over Temporal Networks: 
The Role of Heterogeneities}% Force line breaks with \\

\author{Arash~Badie-Modiri, Chiara~Boldrini, Lorenzo~Valerio, János~Kertész and Márton~Karsai%
\thanks{Arash~Badie-Modiri, Chiara~Boldrini and Lorenzo~Valerio are with the National Research Council, Pisa}%
\thanks{Arash~Badie-Modiri, János~Kertész and Márton~Karsai are with the Central European University, Vienna}%
\thanks{Arash~Badie-Modiri is with Aalto University, Espoo}%
\thanks{Márton~Karsai is with the HUN-REN R\'enyi Institute of Mathematics, Budapest}%
}

\maketitle

\begin{abstract}
Decentralised federated learning, based on peer-to-peer communication, is increasingly proposed for on-device training of machine learning models, promising a privacy-preserving, communication-efficient training process with no risk of single-point failure. However, the role of structural and temporal inhomogeneities in such fully decentralised settings remains poorly understood. Here, we investigate their effects when model parameters are locally averaged during aggregation.  We show that the decentralised federated learning process is governed, both in the early phase and the late, stationary limit, by the same dynamics as a lazy random-walk diffusion process on temporal networks. Based on this mapping, we demonstrate that the typical experimental scenario used in decentralised federated learning leads to unrealistically rapid convergence because of ignoring the temporal and structural inhomogeneities inherent in the communication network. We analyse real-world temporal networks and find that inhomogeneities most often dramatically slow down diffusion, hence the convergence process.
\end{abstract}

\section{Introduction}\label{sec:introduction}

Decentralised federated learning has emerged as a powerful paradigm for training machine learning models across entities without sharing raw data or requiring central coordination \cite{beltran2023decentralized}. This approach is often studied in the context of data centres, for example in healthcare settings where multiple centres contribute patient data to jointly train a model without directly sharing sensitive information  \cite{lian2022deep,wang2022efficient}. However, decentralised federated learning can also be highly beneficial for end-user devices such as smartphones and internet-of-things devices \cite{wang2022accelerating} or autonomous vehicles \cite{chen2021bdfl}, enabling them to contribute to model training using their local data without sharing raw and private data. Unlike the data-centre setting (where robust networking infrastructure allows reliable, persistent, and high-bandwidth connections), edge devices communicate opportunistically: they go offline unpredictably, move through space, and may interact only when physical proximity or shared infrastructure allows. These constraints make the structure and timing of communications inherently irregular and heterogeneous.

Early works addressed this setting under idealised assumptions of synchronous, peer-to-peer communication over simple network topologies~\cite{lian2017can, tang2018d2}. While more recent studies have begun to relax these assumptions by incorporating aspects of structural and temporal heterogeneity into the analysis \cite{palmieri2025robustness, palmieri2024impact, zhai2025decentralized, badie2024initialisation}, the forms considered remain narrow in scope. These works highlight the impact of features such as heterogeneous degree distributions, community structure, and intermittent communication failures on the convergence behaviour of decentralised learning systems. However, prior work largely treats heterogeneity as static topology variation (irregular but fixed graphs) or as one-shot randomly induced disruption, such as links or nodes deactivated independently at random. Neither of these captures the richer dynamics present in real-world temporal networks, where communication patterns may be bursty, temporally correlated, and shaped by memory-bearing processes that couple structure and timing in non-trivial ways.

In parallel, the network science literature has developed a rich theoretical understanding of dynamical processes on complex networks, including diffusion and spreading on systems with intricate structural and temporal patterns (see~\cite{BarratBarthelemyVespignani2012}). While these models provide powerful tools for analysing how information propagates in realistic networks, they are typically studied independently of specific applications such as decentralised learning. Bridging this gap requires connecting learning dynamics with the well-understood behaviour of diffusion processes on temporal networks.

In this work, we establish such a connection by modelling decentralised federated learning as a diffusion process over a temporal communication network. We analyse how local model parameters propagate through time-varying connectivity, capturing both asynchronous communication and complex temporal heterogeneity within a unified framework. We do this through two complementary analyses: (i) in the early synchronisation phase, where aggregation dominates local learning, we characterise how quickly models converge through the lens of random walks on temporal networks; (ii) and in the stationary regime, where we model the spread of perturbations introduced by local training steps as independently diffusing impulses governed by the same diffusion dynamics. We validate this framework on both synthetic networks (introducing spatial embedding, bursty renewal processes, and self-exciting temporal dynamics in isolation) and on real-world contact networks drawn from three distinct deployment scenarios, using microcanonical randomised reference models~\cite{gauvin2022randomized} to isolate the contribution of specific heterogeneity classes.

Our analysis yields three key findings:
\begin{itemize}
\item The dissemination of learned parameters is governed, both in the early and stationary phases, by the same dynamics as a lazy random-walk diffusion process on the underlying temporal network, providing a unified and analytically tractable characterisation of decentralised federated learning convergence in both regimes.

\item Both structural and temporal heterogeneities universally slow down this diffusion process: lower-dimensional spatial embedding, heavier-tailed inter-event time distributions, and stronger self-excitation all push convergence toward slower regimes, and their effects compound in real-world networks where multiple heterogeneities co-occur.

\item The standard experimental setup used in decentralised federated learning research (random graphs with regular communication intervals) is systematically biased toward unrealistically fast convergence; in the real-world networks we study, a fully randomised baseline with the same number of nodes, links, and total events can mix tens to more than a hundred times faster than the empirical network.
\end{itemize}
Together, these results show that network heterogeneity is not a second-order modelling detail but a primary determinant of convergence speed, with direct implications for system design, protocol evaluation, and the interpretation of simulation-based benchmarks.

The remainder of this paper is organised as follows. Section~\ref{sec:related_works} reviews related work on decentralised federated learning and dynamical processes on temporal networks. Section~\ref{sec:system_model} introduces the system model, formalising the decentralised federated learning setup and the temporal network framework. Section~\ref{sec:diffusion-theoretical} establishes the connection between decentralised federated learning and lazy random-walk diffusion, covering both the early-stage synchronisation dynamics and the stationary-phase response to local learning perturbations. Section~\ref{sec:results} presents our experimental results on synthetic networks and real-world networks, quantifying the effect of structural and temporal heterogeneities on convergence. Finally, Section~\ref{sec:conclusion} discusses the implications of our findings and directions for future work.

\section{Related works}
\label{sec:related_works}

Decentralised federated learning is defined on real-world communication networks which often simultaneously embody many types of temporal and structural heterogeneities \cite{holme2012temporal}. Research into temporal networks has shown that simply taking into account timing and order of contacts between nodes, as opposed to using an aggregated static network as a substitute, can strongly affect the rapidity of spreading processes \cite{karsai2014time, karsai2011small}. This has also been shown to hold true for diffusion processes, such as random walks, where unlike in epidemic spreading process, the number of ``spreading agents'' remains conserved \cite{starnini2012random}.
Temporal heterogeneities, often manifesting as bursty and/or correlated communication dynamics and heterogeneity in overall frequency of activation, are often present in real-world systems \cite{karsai2018bursty, barabasi2005origin}. To understand the behaviour of these systems, we have to uncover the role of temporal inhomogeneities in their dynamics \cite{karsai2011small, masuda2013temporal}.

Similarly, a large body of the literature on complex networks has been devoted to understanding the role of structural heterogeneities. Heterogeneous degree-sequence \cite{barabasi1999emergence, newman2001random}, preferential attachment \cite{dorogovtsev2000structure} and small-world property \cite{watts1998collective, milgram1967small} of networks have been at the centre of attention of network scientists for decades. Other works have studied the effects of spatial limitations on networks, showing that the structure and connectivity in many real-world networks are affected by limitations imposed by the fact that these systems are embedded in a finite-dimensional space \cite{barthelemy2011spatial}. This research has been complemented through analyses of the roles of mesoscopic structures in the network, for example motifs (over-represented local patterns of connectivity) \cite{milo2002network, alon2007network} and community structure \cite{newman2004detecting, girvan2002community}. The structural and temporal heterogeneities can combine. For example, temporal motifs capture over-represented \emph{temporal and structural} patterns of connectivity that cannot necessarily be detected on a static aggregation of the network \cite{kovanen2011temporal}.

We now turn to overview the related literature on Decentralized Federated Learning (DFL), which replaces the client--server architecture of classical FL with peer-to-peer communication, where clients alternate local training with neighbourhood model mixing over a graph. Algorithmically, DFL is closely related to decentralized stochastic optimization: convergence couples stochastic gradients with the connectivity and spectral properties of the mixing operator. Canonical baselines such as D-PSGD~\cite{lian2017can} and D$^2$~\cite{tang2018d2} formalize this coupling, and subsequent work studies communication constraints via compressed exchanges~\cite{koloskova2019decentralized}. Empirical and methodological studies further indicate that even under static graphs, structural heterogeneity (e.g., bottlenecks, modularity) can dominate early alignment and affect robustness and final accuracy~\cite{palmieri2024impact, badie2024initialisation}, motivating topology-aware evaluation and coordination-free protocols~\cite{valerio2023coordination} as well as disruption-oriented analyses~\cite{palmieri2025robustness}; surveys summarize protocol choices and open challenges~\cite{beltran2023decentralized}.

The mixing step has deep roots in control and distributed computation. Average-consensus theory relates agreement rates to spectral properties of averaging weights~\cite{OlfatiSaber2007Consensus}, and randomized pairwise exchanges used in DFL coincide with classical gossip schemes~\cite{Boyd2006Gossip}. Switching-topology consensus establishes conditions (e.g., joint connectivity over time windows) guaranteeing agreement under time-varying interactions via products of stochastic matrices~\cite{Jadbabaie2003NearestNeighbor,Moreau2005TimeVaryingLinks}. Related distributed optimization methods explicitly separate optimization and disagreement errors~\cite{Nedic2009DistributedSubgradient}; however, their regularity assumptions (bounded delays, sufficiently frequent exchanges, well-behaved switching) may not hold in empirical contact traces.

Within DFL, several works address communication heterogeneity and time variation at the protocol level. Hu \emph{et al.} propose segmented gossip to better exploit heterogeneous link capacities under serverless training~\cite{Hu2019SegmentedGossipDFL}, while Heged\H{u}s \emph{et al.} empirically show that gossip learning can be competitive with centralized FL across regimes~\cite{Hegedus2019GossipLearning}. For mobility-driven variation, Lu \emph{et al.} study privacy-preserving DFL over time-varying graphs via Metropolis--Hastings weights and secret sharing~\cite{Lu2022PrivacyTemporalDFL}. In wireless settings, Jeong \emph{et al.} propose asynchronous decentralized learning robust to failures~\cite{Jeong2022AsyncWirelessDFL}, and Nguyen \emph{et al.} analyse time-varying \emph{directed} networks using gradient tracking and momentum (DSGTm-TV) with row/column-stochastic mixing~\cite{Nguyen2024DSGTmTV}. Beyond ``DFL over a given network,'' Zhang \emph{et al.} design time-varying mixing matrices for energy-efficient DFL~\cite{Zhang2025TimeVaryingMixing}, and Li \emph{et al.} target time-varying and heterogeneous mobile computing networks~\cite{LiTMCtimeVaryingMobileDFL}.

The present manuscript provides a complementary network-science viewpoint: it models pairwise DFL mixing as diffusion on temporal networks and links convergence-limiting behaviour to temporal mixing and localization. Building on the observation that early synchronization in static graphs is governed by lazy random-walk mixing~\cite{badie2024initialisation}, the paper generalizes to asynchronous contact sequences and proposes the decay of the inverse participation ratio (IPR) as a diagnostic of relaxation and localization in temporal diffusion. This frame of reference explains why common evaluation sets (regular communications in homogeneous random graphs) can overestimate convergence by suppressing structural and temporal heterogeneities observed in real traces, and motivates topology-/time-adaptive communication policies in heterogeneous temporal environments.

\section{System model}
\label{sec:system_model}

In this work, we focus on a simple decentralised federated learning setup, where nodes communicate one-to-one and instantaneously over a dyadic temporal network (contact network) on a continuous-time axis. The temporal network is defined as $G=(\mathcal{V}, \mathcal{E}, \mathcal{T})$, where $\mathcal{V}$  is the set of nodes, $\mathcal{E}$ the set of events and $\mathcal{T}$ the time window of the measurement of the network. Each event $e_m \in \mathcal{E}$ is defined as undirected communication between a pair of nodes at a specific time, $e_m = (\{i, j\}, t_m)$, during which the two nodes exchange their current model parameters (weights and biases) and aggregate their own parameters with those of their peers.

Let each node $i\in\mathcal{V}$ hold parameters\footnote{While for the empirical simulations in this paper we elected to use a variant of the parameter initialisation method described in Ref.~\cite{badie2024initialisation}, the paper is generally agnostic to the choice of the initialisation method.} $w_i(t)\in\mathbb{R}^d$. The dynamics at each node consist of alternating local learning updates and communication-induced aggregation events. Local learning is modelled as (possibly stochastic) updates applied at node-specific times, written abstractly as $w_i \leftarrow \mathcal{U}_i(w_i)$, e.g., one Stochastic Gradient Descent (SGD) step $\mathcal{U}_i(w_i)=w-\eta\,g_i(w_i;\xi)$ with $\mathbb{E}[g_i(w;\xi)]=\nabla f_i(w)$ for a local objective $f_i$. This operator covers any local training rule. Each node internally optimises its parameters using its \textit{local data only}, at a constant rate through the local learning process. 
A communication event $e_m$ couples the states of two nodes through an instantaneous aggregation map $\mathcal{A}$ applied to their pre-event parameters,
\begin{equation}
(w_i,w_j)\leftarrow\big(\mathcal{A}(w_i,w_j),\,\mathcal{A}(w_j,w_i)\big)\,.
\end{equation}
Note that, in this paper, we assume that all events are dyadic and no two events involving the same node occur at the same time. Effectively, we assume that communications involving the same node happening simultaneously are resolved one at a time in random order, although we briefly discuss the ramifications of simultaneous communications between one node and multiple neighbours in \cref{sec:early-stage}. Therefore, unless otherwise specified, we use simple pairwise averaging as our aggregation rule, i.e.,
\begin{equation}\label{eq:pairwise-avg}
w_i \leftarrow \tfrac12(w_i+w_j),\qquad
w_j \leftarrow \tfrac12(w_i+w_j)\,.
\end{equation}
This symmetric aggregation is a widely used baseline. The model can be readily generalized to asymmetric mixing rules, whereas accommodating heterogeneous architectures, non-Euclidean parametrizations, or adaptive mixing would require more substantial modifications. We briefly discuss possible generalisation of our findings to more involved non-linear aggregation processes in \cref{sec:perturbation-response}.

Depending on the phenomenon and the setting under study, temporal networks are sometimes modelled as interval graphs, where a link, once established between two nodes, remains open for a certain amount of time before closing. During this ``on'' interval, communication can occur at the discretion of the nodes. In our target scenario described above, this would be the most realistic modelling of contacts, for example, if two devices stay in proximity for a significant duration. 
% This would be the case, for example, if two devices stay in proximity for a significant duration. 
However, for mathematical tractability, in our theoretical analyses (\cref{sec:diffusion-theoretical}) we did not consider this case, i.e., we assume no duration. In contrast, in the experimental setting (\cref{sec:results}) we approximate prolonged contacts by having nodes exchange messages (``ping'' each other) at fixed intervals, so that after a specified time since the last interaction, a new instantaneous contact event is triggered, effectively converting the interval network into a sequence of instantaneous events. 

Under the instantaneous contact assumption, we effectively assume unbounded communication bandwidth, such that model parameters can be exchanged without delay. This is a deliberate simplification adopted for analytical tractability. We also neglect medium contention, implicitly assuming that simultaneous contacts do not interfere with each other or require scheduling. In practice, however, the volume of information that can be transmitted during a contact depends on the underlying communication technology (e.g., Bluetooth, Wi-Fi) and its associated bandwidth constraints. Incorporating finite, neighbourhood-dependent bandwidth effects into the model is left for future work.

\section{Decentralised federated learning as a diffusion process}\label{sec:diffusion-theoretical}

In this section, we establish the connection between the dynamics of decentralised federated learning and a diffusion process on temporal networks. Previous work has already demonstrated a link between the early-stage ``synchronisation'' phase of decentralised federated learning and lazy random walks, under the restrictive assumptions of a static communication graph, discrete time, and synchronous updates in which all nodes communicate simultaneously at each time step~\cite{badie2024initialisation}. We generalise this connection in two directions: in \cref{sec:early-stage} we extend the analysis to asynchronous communication on a continuous time axis, focusing on the early-stage regime where aggregation dominates local learning; in \cref{sec:perturbation-response} we shift to the stationary regime, modelling local learning as a perturbation and showing that its spread through the network is governed by the same lazy random-walk diffusion operator.

\vspace{-7pt}
\subsection{Lazy passive random walk and early-stage dynamics of decentralised federated learning}\label{sec:early-stage}

We focus on the early-stage dynamics and track how each node's initial parameters (weights and biases at $t_\text{init.}=0$) influence the parameters of all nodes at later times $t$. In this regime, the aggregation step dominates local learning in magnitude, so the learning contribution is negligible \cite{badie2024initialisation}. This lets us focus on the effect of aggregation alone.

\begin{figure}
    \centering
    \includegraphics[width=0.85\linewidth]{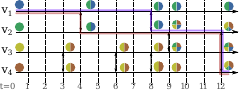}
    \caption{The composition of the parameters of each model under simple average aggregation based on initial node parameters, disregarding changes due to training. The parameters of node $V_4$ just after the final tick $t = 12$ displayed in the schematic is a linear composition of the initial parameters of all nodes, with $1/8$ of it originating from the initial parameters of node $V_1$. This share of influence can be calculated by drawing all possible reverse time-respecting paths $p$ from $V_4$ just after $t=12$ to $V_1$ at $t=0$, shown as the lavender and magenta trajectories, and assigning each one a value of $2^{-inc(p)}$, where $inc(p)$ is the number of events incident to the path $p$, representing the opportunities for dilution along that path. In this case both paths get a value of $1/16$, adding up to a total of $1/8$.}
    \label{fig:diffusion-schematic}
\end{figure}

Consider the example contact network in \cref{fig:diffusion-schematic}. Before the very first contact, the parameters of node $v_1$ are influenced only by its own initial values $w^\text{init.}_1$. When the first contact occurs between $v_1$ and $v_2$ at time $t=4$, pairwise averaging splits the influence equally, so the parameters at $v_1$ become $w^{(t=4)}_1 = \tfrac12 w^\text{init.}_1 + \tfrac12 w^\text{init.}_2$.

More generally, the influence of $w^\text{init.}_i$ on $w^{(t)}_j$ is boosted by each possible time-respecting path (sequence of contacts whose timestamps are strictly increasing and each consecutive pair of contacts share at least one node, see \cite{holme2012temporal}) between $v_j$ at time $t$ and $v_i$ at time $t_\text{init.} = 0$. The actual contribution of that path depends on how many events are incident to the path at any point between 0 and $t$, since each incident event dilutes this contribution by a 1/2. An event is incident to a time-respecting path if at least one end of the event coincides with the time-respecting path. If we denote the contribution of each node's initial condition on $w^{(t)}_j$ as
\begin{equation}\label{eq:params-at-t}
    w^{(t)}_j = \sum_i a_{ij}(t) w^\text{init.}_i\,,
\end{equation}
we can write $a_{ij}(t)$ as
\begin{equation}\label{eq:contributions}
    a_{ij}(t) = \sum_{\mathclap{p \in P_{v_i(0) \rightarrow v_j(t)}}} 2^{-|inc(p)|}\,,
\end{equation}
where $P_{v_i(0) \rightarrow v_j(t)}$ is the set of all time-respecting paths from $v_i$ at time 0 to $v_j$ at time t, and $inc(p)$ denotes the set of all incident events to the path, including the events traversed by the path $p$ as well as those merely branching off the path $p$.

This combinatorial expression has a natural probabilistic interpretation. Consider a lazy random-walk process starting at node $v_j$ at time $t$, running in the reverse direction of the arrow of time. As the random walker arrives at each event, it traverses that event with probability 50\%, or stays in the node it currently occupies. The probability of the random-walk agent arriving at $v_i$ at time 0 is described exactly as \cref{eq:contributions}, with each possible path contributing to the total probability based on the number of ``crossroads'' encountered on that path. The above reasoning is valid for continuously distributed event times.

This shows that the early-stage mixing dynamics in this system are isomorphic to those of a lazy random-walk process. To quantify mixing, we study the variance of a given parameter across different nodes \cite{badie2024initialisation}: lower variance indicates better mixing, hence stronger agreement among nodes' parameters. From \cref{eq:params-at-t}, assuming initial models' parameters are independently drawn across nodes, we can arrive at
\begin{equation}
    \sigma^2(w^{(t)}_j) = \sigma^2(w^\text{init.}_i) \sum_i a^2_{ij}(t)\,.
\end{equation}
As values of $a_{ij}(t)$ are simply the visit probabilities of a (time-reversed) random walk process starting from a node $j$ at time $t$ ending at $i$ at time $0$, the sum $\sum_i a^2_{ij}(t)$ corresponds exactly to the \emph{inverse participation ratio} (IPR) at time $t$ for the time-reversed system. Inverse participation ratio measures how concentrated the probability mass remains on a small subset of nodes. Equivalently, it can be interpreted as the collision probability that two independent copies of the same time-reversed walk, both started from $(j, t)$, end at the same node at time $0$. Its reciprocal, $1/\sum_i a_{ij}^2(t)$, can be understood as the ``effective number'' of initial nodes contributing appreciably to $w_j^{(t)}$.

The inverse participation ratio is a well-established measure of localisation in the study of disordered systems~\cite{evers2008anderson} and has been used to characterise delocalisation in random walks on networks~\cite{paparo2013quantum}. Its behaviour here is intuitive: at $t = 0$, when the walker is certainly at its starting node, the IPR equals $1$; as the walk delocalises, the IPR decays toward its minimum value of $1/N$, attained only when the walker is uniformly distributed across all $N$ nodes. In general, as $t \to \infty$, the IPR approaches $\sum_i \pi_i^2$, where $\pi_i$ is the stationary distribution of the walk, recovering $1/N$ in the special case of a uniform stationary distribution. Thus, lower IPR values indicate stronger delocalisation, while higher values indicate stronger localisation.

Throughout this paper, we use the rapidity of this decay (the trajectory of the inverse participation ratio as a function of time) as a measure of the delocalisation or ``relaxation'' of the diffusion process across different networks. Because parameter variance across nodes is proportional to the inverse participation ratio, faster decay corresponds directly to faster mixing in the decentralised learning system.

Generalising the same idea for the case where one node communicates simultaneously with $k$ neighbours is straightforward: in this scenario the equivalent random walk process stays on the node with probability $1/(k+1)$. The generalisation of \cref{eq:contributions} for this scenario, describing both the random-walk process as well as the influence of $w^\text{init.}_i$ on $w^{(t)}_j$, is
\begin{equation}
a_{ij}(t) = \sum_{\mathclap{p \in P_{v_i(0) \rightarrow v_j(t)}}} \prod\nolimits_{e \in inc(p)} |e|^{-1}\,,    
\end{equation}
where $|e|$ is the number of nodes participating in event $e$, i.e., $k + 1$ in the above example. This process, while distinct from the typical definition of lazy random walk, shares important characteristics with lazy random walks. For example, since $0 < 1/(k+1) < 1$, the mixing time of this process asymptotically grows with the typical lazy random walk up to a constant factor~\cite[Corollary 9.5]{peres2015mixing}. For simplicity, however, the remainder of this manuscript will only consider one-to-one, instantaneous communications on a continuous axis of time, assuming that possible simultaneous events are resolved one at a time in a random order.

\vspace{-7pt}
\subsection{Stationary dynamics and the response to perturbations}\label{sec:perturbation-response}

The analysis in \cref{sec:early-stage} characterises the early-stage dynamics, where the aggregation process dominates and local learning contributes negligibly to parameter evolution. As the system approaches stationarity, the parameters of different nodes have largely aligned, and the dominant source of variation becomes the ongoing local learning process at individual nodes. Analysing the effect of local learning directly is difficult, as the learning operator is in general non-linear: a training step at node $i$ depends on the current parameter vector $w_i(t)$ through the local loss landscape, making learning-induced changes path-dependent and typically correlated across nodes. However, once a learning-induced perturbation $\delta w_i$ is injected at a node, its subsequent redistribution through the network is governed by the aggregation process alone. And aggregation, being a linear operation, means that the response to any collection of perturbations is simply the superposition of single-impulse responses. 

Our diffusion characterisation should therefore be read as a statement about \emph{propagation}, and not about \emph{generation}, of updates. Once a learning-induced perturbation $\delta w_i$ is injected at a node, regardless of its direction and magnitude, its subsequent redistribution through the network is governed by the same linear diffusion operator as in the averaging-only dynamics. In particular, conditional on a realised sequence of learning perturbations, their contributions to $w(t)$ superpose linearly. Each increment spreads according to the same (time-dependent) diffusive relaxation kernel, and the total deviation is the sum of the diffused increments. Non-linear effects enter only through the fact that the increments themselves depend on the evolving state, and are therefore neither independent nor identically distributed.

In an infinitesimally short time period $\mathrm{d} t$, the parameters of node $i$ evolve as
\begin{multline}
    E[w_i^{(t+dt)} - w_i^{(t)}] = - \frac{1}{2} \sum_j \lambda_{ij}(t) \left(w_j^{(t)}  - w_i^{(t)}\right) \mathrm{d} t \\+ o(\mathrm{d} t)\,,
\end{multline}
where $\lambda_{ij}(t)$ is the instantaneous interaction rate between $i$ and $j$. This expression has a natural interpretation: at rate $\lambda_{ij}(t)$, node $i$ averages with node $j$, which pulls $w_i$ toward $w_j$. The expected drift is therefore proportional to the disagreement $w_i^{(t)} - w_j^{(t)}$, summed over all neighbours $j$ and scaled by the interaction rates. To express the dynamics compactly, we write $w^{(t)}=(w^{(t)}_1,\dots,w^{(t)}_n)$ for the stacked parameter vector across all nodes, and define $L_{ij}=(e_i-e_j)(e_i-e_j)^\top$, where $e_i$ is the $i$-th standard basis vector. The weighted Laplacian of the network is then $L_\lambda(t)=\sum_{i<j}\lambda_{ij}(t)L_{ij}$. With this notation, the expected infinitesimal evolution of the full system becomes
\begin{multline}\label{eq:kernel-update}
    E[w^{(t+dt)} - w^{(t)}] = -\frac{1}{2} L_\lambda(t) w^{(t)} \mathrm{d} t +  o(\mathrm{d} t).
\end{multline}
Thus, the propagation of perturbations is governed by the linear operator $-\frac{1}{2} L_\lambda(t)$, which is precisely the diffusive relaxation operator associated with a lazy continuous-time random walk (i.e., a process that, at each interaction opportunity, moves with probability $1/2$ and stays put otherwise). In this sense, parameter deviations diffuse across the network according to the same Laplacian dynamics as the corresponding random walk.

\begin{figure}
    \centering
    \includegraphics[width=0.8\linewidth]{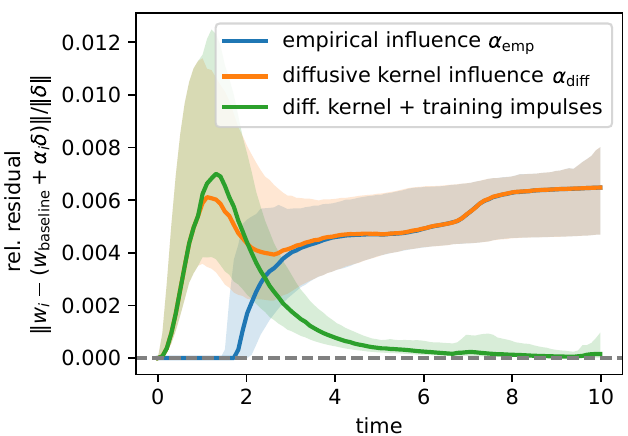}
    \caption{Prediction accuracy for the parameters of each node after a perturbation of magnitude $\|\delta\|$ in a decentralised federated setting using the diffusive kernel of the underlying temporal network (orange) quickly converges to the prediction based on the the entire history of node contacts (blue), but both predictions start diverging from reality because of the effect of local training on the nodes. If we also take into account each successful local training event as an additional impulse, and sum up the total effects additively using the diffusive kernel method (green) the diffusive kernel can predict the trajectory of all network nodes even in fairly long time periods. Note that the initial peak in the diffusive kernel trajectories (orange and green) is due to the random differences between this particular realisation of temporal network. If the experiment is repeated for an ensemble of networks generated from the same parameters, this difference goes away (see Appendix~\ref{sec:perturbation-experiment-apdx} and \cref{fig:influence-empirical}).}
    \label{fig:residual-single}
\end{figure}

\cref{fig:residual-single} shows that the relaxation of a small perturbation of magnitude $\|\delta\|$ on a single node in an otherwise homogenised decentralised federated learning regime is fairly well-predicted by the diffusive kernel of the underlying temporal network. This can be verified by comparing the magnitude of the residual (orange trajectory in \cref{fig:residual-single}, the difference between what diffusive kernel $-\frac{1}{2} L_\lambda(t)$ estimates the parameters of each node should be versus what they actually are) to the magnitude of the initial perturbation $\|\delta\|$. For short time horizons, the residual remains limited to only a fraction of the initial perturbation $\|\delta\|$, indicating that the diffusion-based approximation accurately captures the redistribution of the perturbation.

When the residuals are calculated from empirical influence (blue trajectory), i.e., extracted from the actual temporal network events,  as opposed to the Laplacian (orange trajectory), the initial increase in residual is much more gradual. Over longer time scales, however, both predictions gradually accumulate bias due to ongoing local learning at different nodes. These learning-induced updates introduce additional perturbations that are not accounted for when only the initial impulse is propagated.

If, instead, we model both the initial perturbation and the subsequent local training updates as independent impulses that each relax according to the same diffusive kernel $-\frac{1}{2} L_\lambda(t)$, the resulting superposed prediction (green trajectory in \cref{fig:residual-single}) closely matches the observed dynamics over substantially longer time periods. This is done by calculating the effect of each local training update as well as the initial perturbation independently, then linearly combining the expected effect for each node at each time. For the details of this experiment, including the experimental setup, we refer to Appendix~\ref{sec:perturbation-experiment-apdx}.

The derivation can be trivially extended for the case of aggregation of more than two nodes at the same time, corresponding to lazy random walk on a temporal hypergraph. In this case, the probability of remaining at a node at event $e$ corresponds to $1/|e|$ where $|e|$ is the order (number of incident vertices) of event $e$. For the sake of simplicity, in this paper we only performed experiments on dyadic temporal networks.

The linear relaxation for small perturbations is valid also for a class of non-linear aggregation functions. For example, the \verb|DecDiff| aggregation method \cite{valerio2023coordination} with the update rule
\begin{equation}
w_i^{(t)}=w_i^{(t-1)}+\frac{\bar{w}_i^{(t-1)}-w_i^{(t-1)}}
{\left\lVert \bar{w}_i^{(t-1)}-w_i^{(t-1)} \right\rVert_2+s}\,,
\end{equation}
where $1 \leq s < \infty$ is a hyper-parameter and $\bar{w}_i$ represents the average model in an interaction. Assuming the parameters of node $i$ and its (potential) neighbours diminish in comparison to the constant $s$, i.e., $\left\lVert \bar{w}_i^{(t-1)}-w_i^{(t-1)} \right\rVert_2 \ll s$, this aggregation method can be expanded into a linear operation. Let $d_i^{(t-1)} := \bar{w}_i^{(t-1)} - w_i^{(t-1)}$. Then the \verb|DecDiff| update reads
\begin{equation}
w_i^{(t)} = w_i^{(t-1)} + \frac{d_i^{(t-1)}}{\lVert d_i^{(t-1)} \rVert_2 + s}\,.
\end{equation}
If $\lVert d_i^{(t-1)} \rVert_2 \ll s$, we expand
\begin{equation}
\frac{1}{\lVert d \rVert_2 + s} = \frac{1}{s}\Big(1 + \frac{\lVert d \rVert_2}{s}\Big)^{-1}
= \frac{1}{s} + \mathcal{O}\!\left(\frac{\lVert d \rVert_2}{s^2}\right),
\end{equation}
which yields the first-order linear approximation
\begin{align}\label{eq:decdiff-linearised}
w_i^{(t)} 
=& \Big(1-\frac{1}{s}\Big) w_i^{(t-1)} + \frac{1}{s}\,\bar{w}_i^{(t-1)} + \mathcal{O}\!\left(\frac{\lVert d_i^{(t-1)} \rVert_2^2}{s^2}\right).
\end{align}
Thus, in the small-discrepancy regime \verb|DecDiff| reduces to a weighted linear averaging toward the interaction average $\bar{w}_i$, and its propagation through the network is governed by the same diffusion operator as pairwise averaging, up to a constant rescaling. More broadly, any aggregation rule that reduces to a weighted linear average in the small-discrepancy regime will inherit the same diffusive propagation dynamics, up to a constant rescaling of the effective interaction rate.

\section{Results}\label{sec:results}
Building on the characterisation established in \cref{sec:diffusion-theoretical}, we now investigate how structural and temporal heterogeneities in the communication network affect the rapidity of the diffusion process, and by extension the convergence behaviour of decentralised federated learning. Most experimental and analytical works in this area assume that nodes communicate at regular intervals over homogeneous network topologies,  either regular structures such as ring networks~\cite{wang2022efficient} or realisations of simple random models such as the Erdős–Rényi model~\cite{valerio2023coordination, palmieri2024impact}. More recent works have begun to incorporate degree heterogeneity~\cite{palmieri2024impact} or induce temporal heterogeneity through random link or node deactivation~\cite{zhai2025decentralized, badie2024initialisation}, but these remain limited in scope. We show that these common assumptions cause a significant and systematic underestimation of the time required for the diffusion process to relax, which in turn governs convergence in decentralised federated learning.

We proceed in two steps. First, in \cref{sec:structural,sec:temporal} we use synthetic generative models to isolate the effect of specific classes of heterogeneity (spatial embedding, burstiness, and self-excitation) introduced one at a time into an otherwise homogeneous baseline. Second, in \cref{sec:results_traces} we move to real-world contact networks, where multiple heterogeneities co-occur, and use microcanonical randomised reference models~\cite{gauvin2022randomized} to disentangle their individual contributions. In both settings, convergence is measured through the decay of the IPR $\sum_i P_i^2(t) - 1/N$, with faster decay corresponding directly to faster mixing (see \cref{sec:early-stage}).

Our results show that heterogeneities universally slow down the diffusion process, in both synthetic and real-world networks. The effect can be dramatic: a fully randomised network with the same number of nodes, links, and total communication events as an empirical network can mix tens to more than a hundred times faster than the original, confirming that standard experimental setups are systematically biased toward unrealistically fast convergence.

\vspace{-10pt}
\subsection{Structural heterogeneities}\label{sec:structural}

\subsubsection{Spatial dimensionality}
\label{sec:results_spatial}

In many different types of real-world networks, the nodes are embedded in some form of geographical or physical space and the formation of links between nodes is affected by the distance between pairs of nodes \cite{barthelemy2011spatial}. For example, communication networks such as device-to-device ad-hoc wireless networks \cite{hekmat2006connectivity, gupta2002capacity} or infrastructure networks \cite{vazquez2002large, buhl2006topological} typically exhibit strong distance–decay in tie formation. Even in systems where spatial constraints are not immediately apparent (such as mobile phone communication), interaction patterns often retain a strong spatial signature due to infrastructure and behavioural constraints \cite{lambiotte2008geographical}. While this embeddedness in finite-dimensional space has shown in the past to significantly affect the rapidity and extent of spreading processes \cite{badie2022directed}, it is often not represented in many commonly used synthetic models (such as the ones used in Sec.~\ref{sec:temporal}, effectively corresponding to a high-dimensional regime where spatial effects vanish).

To evaluate the role of spatial embedding on the early-stage dynamics of decentralised federated learning, we use a geometric random graph model. Such models capture interactions constrained by a $d$-dimensional space, such as wireless communication networks and human mobility and interaction networks \cite{barthelemy2011spatial}. The geometric graph model used here is parametrised by number of nodes $N$, average degree $\langle k \rangle$, system dimensions $d$ and temperature $T$. We model the network topology as a temperature-controlled random geometric graph on the surface of unit $(d+1)$-dimensional sphere $S^{d+1}$. Nodes are placed i.i.d. uniformly on the surface. The probability of a connection (edge) between two nodes $u$ and $v$, $u \neq v$, is
\begin{equation}
    p_{uv} = \min \left\{ 1, c\theta_{uv}^{-d/T}\right\}\,,
\end{equation}
where $\theta_{uv}$ is great-circle (geodesic) distance between $u$ and $v$, constant $c$ is selected based on a numeric solution so that the resulting graph has the desired expected mean degree $\langle k\rangle$, and $T > 0$ is the temperature parameter that controls the strength of distance dependence. Existence of edges is independent of each other.

A temperature approaching zero can be interpreted similarly to a threshold model: connections to closer nodes are incomparably more likely than connections to nodes further away. As the temperature increases, the difference in probabilities due to distance gets smaller. A geometric random graph model with temperature of infinity shows no effect from dimensionality of the nodes and degenerates to an Erdős–Rényi model. Throughout this paper we use $T = 0$ as a shorthand notation for the above model at the extreme limit of $T$, where it approaches a threshold model in nature. The model interpolates smoothly between realistic low-dimensional spatial networks (small $d$, small $T$) and the Erdős–Rényi ensemble, which is recovered both in the high-dimensional limit $d \to \infty$ and the high-temperature limit $T \to \infty$, where the dependence of $p_{uv}$ on $\theta_{uv}$ vanishes and all node pairs become equally likely to connect. The standard decentralized federated learning experimental setup (homogeneous random graphs) therefore corresponds to the latter end of this spectrum.

\begin{figure}
    \centering
    \includegraphics[width=\linewidth]{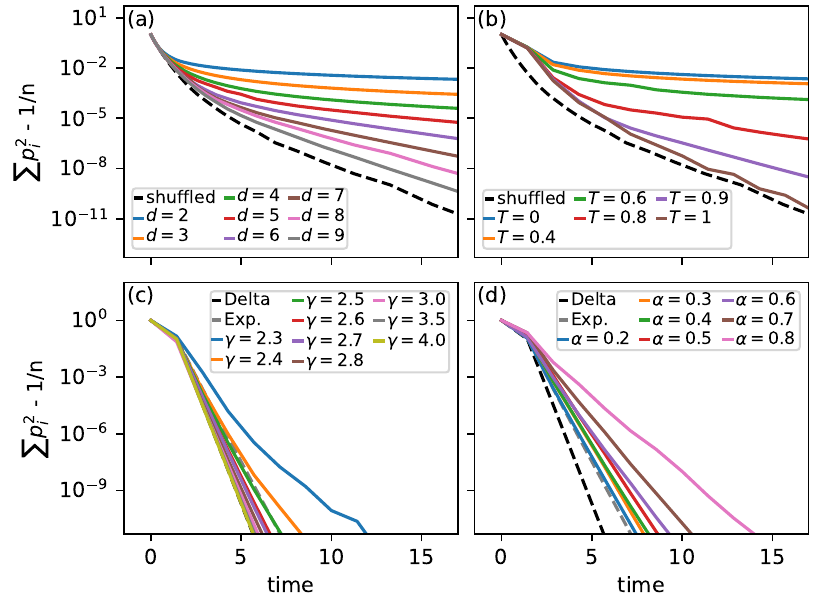}
    \caption{Relaxation of the diffusion process on random network models, governed by the same dynamics as decentralised federated learning, showing that the experimental setups commonly used in decentralised learning studies can yield unrealistically rapid diffusion of information. We analyse the effect of (a) spatial dimensionality, (b) distance-decay temperature, (c) burstiness induced by a renewal process with power-law inter-event times, and (d) self-excitation. In all panels, the black dashed curve denotes the deterministic baseline with constant inter-event times (the $\delta$-distributed case), corresponding to the regular communication schedule often assumed in decentralised learning experiments. In panels (a) and (b), the shuffled baseline refers to a randomized network in which the spatial heterogeneity is removed while preserving the corresponding coarse network constraints, serving as a homogeneous reference representing experimental scenarios with no spatial heterogeneity modelled. In all four experiments, the baseline cases, both spatially shuffled and constant $\delta$-distributed inter-event time, lie at the fast-mixing end of the spectrum. In panels (c) and (d), the grey dashed trajectory denotes the Poisson baseline with exponentially distributed inter-event times.}
    \label{fig:random-model}
\end{figure}

For the synthetic network experiment in this section, we generated ensembles of 30 (or 90 for $d = 9$) geometric static networks with $2^{15}$ nodes and average degree 10. For the dimensionality experiment (\cref{fig:random-model}a) we used a temperature of $T=0$ and for the temperature experiment (\cref{fig:random-model}b) we used dimensionality $d=2$. Events in both experiments are generated using a Poisson process with mean inter-event time of 1.

The results in \cref{fig:random-model}(a) show that the dimensionality significantly affects the relaxation time, with a 
%significant 
considerable difference between two or three dimensional systems, common in the real-world networks with a physical component, and the high-dimensional networks often used in simulated studies. The degree-preserving shuffled baseline (dashed line), which removes spatial structure while preserving node degrees, closely tracks the high-dimensional curves, confirming that it is the spatial embedding, and not the degree sequence, that drives the slowdown. This \emph{high-dimensional} (effectively mean-field) system represents the experimental setup often used in decentralised federated learning simulations in the literature. The slowdown effect of lower dimensionality on diffusion processes matches with previous results on similar phenomena such as spreading processes \cite{moore2024network, badie2022directed, badie2022directed2}.

Another important parameter is the temperature. \Cref{fig:random-model}(b) shows that 
as the temperature decreases toward the threshold model, the effect of low dimensionality is increasingly accentuated: stricter geometric constraints produce slower diffusion, while higher temperatures progressively wash out the spatial structure and recover the fast-mixing Erdős–Rényi behaviour.

\vspace{-10pt}
\subsection{Temporal heterogeneities}\label{sec:temporal}

While structural heterogeneities, studied in Section~\ref{sec:structural}, are fixed properties of the communication graph, temporal heterogeneities arise independently of topology: the same static network can exhibit vastly different diffusion behaviour depending on the timing of its events. We study two canonical mechanisms (heavy-tailed inter-event times and self-excitation), which together capture the most commonly observed forms of temporal irregularity in real-world contact networks. In both cases we use the same underlying spatial network: ensembles of 30 geometric static networks with $2^{15}$ nodes, $d = 2$ dimensions, $T = 0$ temperature and average degree 10, with event times parametrised to produce a mean inter-event time of 1. This holds the static topology fixed across both experiments so that any observed differences in diffusion rapidity are attributable to temporal structure alone.

\subsubsection{Bursty renewal-processes}
\label{sec:results_bursty}

Burstiness refers to the observation that events in many real-world temporal networks tend to cluster in short time intervals followed by relatively long periods of inactivity. This is in contrast to the Poissonian assumption that events arrive at a steady constant rate, or the even more simplistic assumption that events happen at precise regular intervals. Burstiness appears in a wide range of systems, from social communication networks to biological interaction patterns \cite{karsai2018bursty}.

Burstiness affects the rapidity of diffusion processes because a flurry of events in a narrow window can rapidly spread information or contagion across a network. At the same time, extended lulls can delay propagation \cite{karsai2011small}. As a result, incorporating burstiness into models of temporal networks is important for accurately capturing how information travels in the systems.

For the purposes of this section, we model burstiness by assuming that consecutive inter-event times are independent draws from a power-law distribution with exponent $\gamma$, that is, the interaction sequence follows a renewal process with heavy-tailed waiting times.

In the experiments here, we vary the power-law exponent $\gamma$. The results in \cref{fig:random-model}(c) show that as burstiness ($\gamma \rightarrow 2$) increases, the rapidity of the relaxation of the diffusion process decreases significantly. In other words, more strongly heavy-tailed inter-event time distributions lead to markedly slower mixing. For comparison, the case of perfectly regular interactions (where inter-event times are deterministic, i.e., follow a delta distribution), shown as the dashed black curve in the figure, represents one of the fastest relaxation regimes. Similarly, large values of $\gamma$, for which the power-law distribution becomes less heavy-tailed and closer to exponential-like behaviour, also lie toward the fast-mixing end of the spectrum.

\subsubsection{Self-exciting processes}
\label{sec:results_selfexciting}

Self-exciting processes are models in which the occurrence of an event between two nodes increases the likelihood of subsequent events happening between the same pair soon after. The renewal processes studied in Section~\ref{sec:results_bursty}, instead, treat consecutive inter-event times as independent draws, so the history of a link's activations carries no information about its future behaviour. Self-exciting processes relax this assumption by introducing memory: each event raises the probability of further events on the same link in the near future, producing temporal clustering through a fundamentally different mechanism.

We model self-excitation using a Hawkes process~\cite{hawkes1971point}, popular in finance, seismology, and social networks,  and specifically a Hawkes univariate exponential process~\cite{laub2021elements}, where each event on a link raises the instantaneous rate of subsequent events on that same link, with influence decaying exponentially over time. The process is parametrised by a background rate $\mu$, which drives spontaneous activations, an infectivity factor $\alpha$, which controls the expected number of additional events triggered by each activation, and a decay rate $\theta$, controlling how quickly the self-exciting influence of each event fades. Constraining $\alpha + \mu = 1$ yields a process with mean inter-event time of 1, where $\alpha$ controls the fraction of events generated by self-excitement and $\mu = 1 - \alpha$ the fraction driven by the constant background rate. The special case $\alpha = 0$ degenerates to a Poisson process, i.e., renewal process with exponential inter-event times, firing at a constant rate of 1. In our experiment, for the sake of simplicity, we used equal values of $\mu$, $\alpha$ across all edges, and fixed $\theta = 1$.

The results in \cref{fig:random-model}(d) show that as a larger fraction $\alpha$ of events are generated as a result of self-excitement and a smaller fraction $\mu = 1 - \alpha$ through random constant background rate, the diffusion process dramatically slows down. In other words, stronger temporal clustering of events leads to slower relaxation. Once again, the deterministic baseline with constant inter-event times (the delta-distributed case, black dashed line) lies at the fastest end of the spectrum. The Poisson baseline with exponentially distributed inter-event times (the $\alpha=0$, $\mu=1$ limit of the Hawkes process, grey dashed line) is slower than the regular-interval case but still faster than all self-exciting cases shown here.

\subsection{Heterogeneities in real-world network}
\label{sec:results_traces}

While it is worthwhile to study the effect of specific types of temporal and spatial heterogeneities in isolation, it is reasonable to assume that the co-occurrence of various heterogeneities might induce effects beyond the sum of their individual contributions. At the same time, these heterogeneities might be correlated in non-trivial ways, for example where structural and temporal dynamics of a node or a group of nodes are both influenced by confounding factors not captured by a temporal network representation of the system.

To disentangle their individual contributions, we start from real-world datasets and apply \textit{microcanonical randomised reference models}~\cite{gauvin2022randomized}: for a given heterogeneity class to be studied, one constructs an ensemble of networks in which that heterogeneity is destroyed by randomisation while all other properties of the system (including its size, density, and the heterogeneities not under study~\cite{karsai2011small}) are preserved exactly in every realisation. By comparing the diffusion behaviour of the original network against such an ensemble, one can isolate the contribution of the targeted heterogeneity class to the observed slowdown.

\subsubsection{Datasets and pre-processing steps}
The three datasets were selected so as to cover qualitatively distinct communication regimes that are representative of plausible deployment environments for decentralised federated learning. For these experiments, we selected (1) a dataset of human proximity from a high school, recorded using RFID devices \cite{fournet2014contact}, (2) a dataset of cab trajectories in San Francisco \cite{piorkowski2022crawdad} and (3) Wi-Fi connection logs from KTH campuses \cite{pajevic2022crawdad}.

The first dataset aims to represent the hypothetical scenario where the ad-hoc communications between devices are governed mostly by the same dynamics as human face-to-face communication. Each node represents a high-school student equipped with an RFID devices, and an event is recorded once every 20 seconds if the RFID devices of two students record a face-to-face interaction. This dataset shows strong diurnal and weekly activity rhythm, bursty dynamics as well as a heterogenous mixing pattern between nodes belonging to different groups, properties often shared with other human social communication networks and proximity networks \cite{barabasi2005origin, malmgren2008poissonian, cattuto2010dynamics}.

The second dataset simulates a scenario where the ad-hoc short-range communications between devices are governed by the same dynamics as opportunistic communications between vehicles moving on a dense urban road network. This is relevant to decentralised learning tasks in which vehicles exchange model updates when they come within communication range. Each event represents a taxi operating in the San Francisco Bay Area. While the original dataset provided location trajectories, we created an event in each 60-second time bin between two vehicles if they come within a 100-meter straight-line distance of each other in that bin.

The final dataset emphasises a mixture of characteristics of human activity and spatially fixed infrastructure. This is relevant to decentralised learning tasks in which movement of devices follows human mobility patterns, but interactions between those devices are mediated with fixed infrastructure such as cell towers or WiFi access points. In this scenario, an event is created between two nodes (students and staff at KTH) once in every hour they are connected to any of the access points of the same building. Since logs do not include sign-offs, each connectivity WiFi log item is taken to indicate twenty minutes of connection to that access point, unless a new log item from the same user is observed.

All datasets were pre-processed as follows: after clipping the events to the desired time window $[t_\text{start}, t_\text{end}]$, a static directed reachability graph was created, where node $i$ is connected to node $j$ if there is a valid, time-respecting path between the two nodes by traversing the events that starts at node $i$ at $t_\text{start}$ arrives at node $j$ before $t_\text{start} + 3/4 (t_\text{end} - t_\text{start})$. The largest strongly connected component of this connectivity graph is used for selecting a well-connected core of the original temporal network where all nodes are eventually reachable from all others. This is done to avoid inclusion of nodes or small subset of nodes where the connection to the rest of the network is tenuous or non-existent, e.g., a vehicle that is not active at all or only active for a brief period before the end of the observation window. Statistics about the resulting processed temporal networks are presented in \cref{tab:datasets}.

\begin{table}[t]
    \centering
    \caption{Datasets used in this work after pre-processing.}
    \begin{tabular}{lrrr}
         \textbf{Dataset} & \textbf{Nodes} & \textbf{Events} & \textbf{Time Window} \\
         \midrule
         SocioPatterns High-School \cite{fournet2014contact} & 178 & 44855 & 202.6 hours \\
         San Francisco Cab \cite{piorkowski2022crawdad} & 508 & 804254 & 55.5 hours \\
         KTH Campus Wi-Fi Log \cite{pajevic2022crawdad} & 4720 & 349087 & 120.0 hours \\
         \bottomrule
    \end{tabular} \vspace{-15pt}
    \label{tab:datasets}
\end{table}

\subsubsection{Randomisation models}\label{sec:randomisation-models}
To identify which empirical heterogeneities limit diffusion, we compare each dataset to ensembles of randomized reference networks in which selected structural or temporal organisation is destroyed while coarse properties of the trace are preserved \cite{gauvin2022randomized,karsai2011small}. The logic is contrastive: if randomising a particular feature accelerates diffusion, that feature was contributing to the slowdown in the original network; the more it accelerates, the larger its contribution. We summarise the set of all types of randomizations considered in Table~\ref{tab:randomisations}, while we refer the interested reader to Appendix~\ref{sec:randomised-reference-models} for the full description of how they are obtained. 

In the structural controls, whole link timelines are reassigned between node pairs. The most conservative variant, Topology-constrained, preserves the exact static projection of the temporal network (i.e., which pairs of nodes have at least one contact) while randomising when and how often they interact. Degree-constrained relaxes this further by preserving only the number of contacts per node, allowing the identity of neighbours to change. The least constrained variant, Link, preserves only the total number of distinct links, randomising both the degree sequence and the contact topology. Progressing from Topology-constrained to Link therefore reveals how much of the structural slowdown is attributable to the fine-grained topology versus the degree sequence alone. In the temporal controls, the static contact graph is held fixed and only the timing of events is shuffled. The most conservative variant, Activity-constrained, preserves the activity window of each link (i.e., the time between its first and last event) while redistributing events uniformly within that window. Weight-constrained additionally preserves the exact number of events on each link. Timeline randomises both the activity window and the number of events, preserving only the total event count across the network. Comparing these variants isolates the contribution of activity window heterogeneity, event frequency heterogeneity, and their combination.

%%% THIS IS JUST A TEST

\begin{table*}[t]
\centering
\caption{Summary of randomised reference models used in this work.}
\label{tab:randomisations}
\begin{tabular}{llp{5.5cm}p{5.5cm}}
\toprule
\textbf{Model} & \textbf{Type} & \textbf{What is preserved} & \textbf{What is randomised} \\
\midrule
Topology-const. & Structural & 
    Static projection (who contacts whom), degree sequence, link weights & 
    Timing and ordering of events on each link \\
\addlinespace
Degree-const. & Structural & 
    Degree sequence, link weights & 
    Contact topology, timing and ordering of events \\
\addlinespace
Link & Structural & 
    Total number of links, link weights & 
    Contact topology, degree sequence, timing and ordering \\
\midrule
Activity-const. & Temporal & 
    Static projection, activity window of each link & 
    Event times within each link's activity window \\
\addlinespace
Weight-const. & Temporal & 
    Static projection, number of events per link & 
    Event times and activity window of each link \\
\addlinespace
Inter-event & Temporal & 
    Static projection, inter-event time distribution per link, time of first event & 
    Order and correlations of inter-event times \\
\addlinespace
Interval & Temporal & 
    Static projection, inter-event time distribution per link & 
    Order, correlations, and start time of each link's timeline \\
\addlinespace
Timeline & Temporal & 
    Static projection, total event count & 
    Activity window, event frequency per link, event times \\
\midrule
Link\,+\,Timeline$^{\dagger}$ & Combined & 
    Total number of links, total event count & 
    Contact topology, degree sequence, all temporal structure \\
\bottomrule
\end{tabular}
\par\smallskip
\scriptsize{$^{\dagger}$ Obtained by sequentially applying Link and Timeline 
shuffling; the resulting ensemble is itself a valid microcanonical reference model. Note that this combination yields the closest analogue to the idealised experimental setup commonly used in the DFL literature, effectively producing an Erd\H{o}s--R\'enyi network with Poisson-distributed event times.}
\end{table*}

\subsubsection{Results on real-world traces vs randomized models}
\begin{figure}
    \centering
    \includegraphics[width=\linewidth]{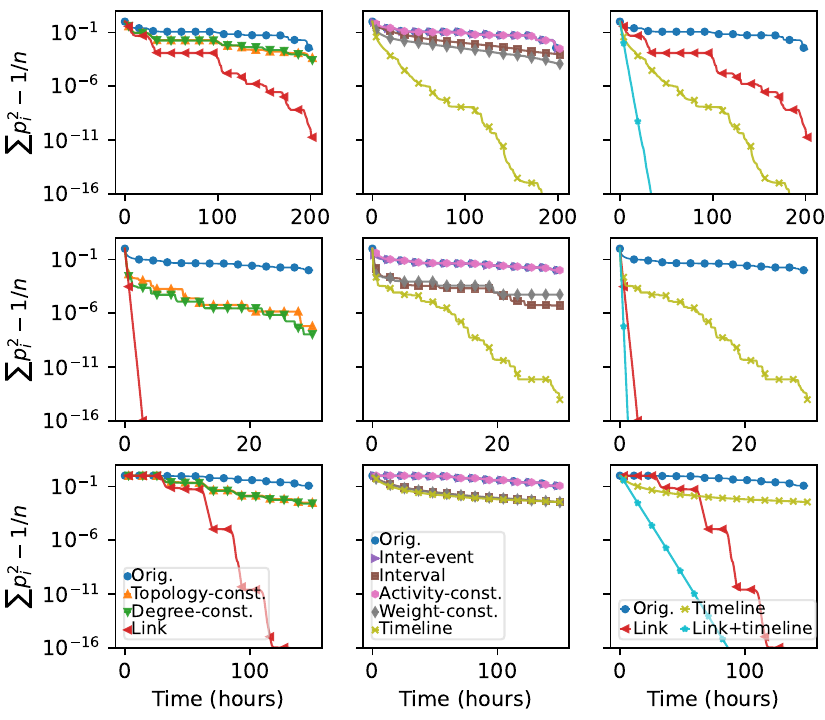}
    \caption{Randomising structural and temporal heterogeneities increases the rapidity of the diffusion process in various real-world temporal networks, as measured by the decay of the inverse participation ratio $\sum_i P^2_i(t)$ signifying the extent of relaxation. Three datasets, high-school face-to-face interactions \cite{fournet2014contact} (a-c), cab trajectories in San Francisco \cite{piorkowski2022crawdad} (d-f) and Wi-Fi connection logs from KTH campuses \cite{pajevic2022crawdad} (g-i), exhibit various forms of structural and temporal heterogeneities. Using microcanonical randomisation methods, we target specific classes of heterogeneities, constructing in each case an ensemble of random networks that preserves large-scale properties of the system (e.g., density and size) as well as the effect of other possible heterogeneities not targeted, as described in \cref{sec:randomisation-models}. The results show that both the structural and temporal heterogeneities inherent in the datasets universally and often dramatically slow down the diffusion process. The typical random network modelling of the system, cyan line in (c, f, i), severely overestimates the rapidity of diffusion compared to the original network, shown as the blue line.}
    \label{fig:real-world}
\end{figure}

Having described the randomisation models, we now examine their effect on the diffusion process across the three datasets.
\Cref{fig:real-world} is organised by perturbation type: rows correspond to the datasets in \cref{tab:datasets}, panels (a,d,g) randomize structure, panels (b,e,h) randomize temporal ordering on the fixed static support, and panels (c,f,i) combine both.

% Structural shufflings separate edge-specific activity, degree heterogeneity, and higher-order topology, while temporal shufflings separate inter-event memory, activity windows, link weights, and global timing. The Link+timeline baseline removes both structural and temporal organisation while preserving only the numbers of nodes, links, and events, making it the closest analogue of homogeneous random communication benchmarks. See \cref{sec:randomised-reference-models} for a full description of the randomised reference models and their use here.

Across all three datasets, randomisation accelerates relaxation in Fig.~\ref{fig:real-world}. In the structural controls (panels a, d, g -- first column of Fig.~\ref{fig:real-world}), the topology- and degree-constrained curves nearly overlap in all three datasets, indicating that preserving the degree heterogeneity (as Degree const. does, see Table~\ref{tab:randomisations}) is sufficient to reproduce most of the structural slowdown. The additional topological detail preserved by the topology-constrained model, such as the specific pattern of links, contributes little beyond what the degree sequence alone already captures. The substantially faster Link baseline, which removes the degree sequence heterogeneity as well, shows that it is primarily the heterogeneity of node degrees that drives the structural slowdown.
In the temporal controls (panels b, e, h -- second column of Fig.~\ref{fig:real-world}), the activity-constrained baseline overlaps almost entirely with the original in all three datasets. This means that preserving the activity windows of each link (i.e., the time between the first and last event on that link) is sufficient to reproduce the original slowdown: the internal timing of events within those windows is irrelevant, and the structure of when each link is active and inactive is what primarily governs temporal diffusion. Interval and weight-constrained shufflings, which additionally randomise the start time of activity windows and the number of events per link respectively, produce a moderate acceleration whose relative ordering varies slightly across datasets, suggesting that these features play an important role. Timeline shuffling, which removes all temporal organisation beyond the total event count, consistently produces the largest acceleration across all three datasets, confirming that the full combination of activity window structure and activation frequency heterogeneity is what most severely limits diffusion in real-world temporal networks.
The combined panels (c, f, i -- third column of Fig.~\ref{fig:real-world}) compare the Link and Timeline baselines, which are the fastest curves in the structural and temporal controls respectively, individually against their sequential combination, Link+timeline. In all three datasets, the Link+timeline baseline is faster than either Link or Timeline alone, showing that structural and temporal heterogeneities contribute independently to the slowdown and that removing both together produces a greater acceleration than removing either one separately. The gap between the original network and the Link+timeline baseline spans several orders of magnitude in IPR across all three datasets, providing a concrete quantification of how severely standard benchmarks, which effectively correspond to this fully randomised regime, overestimate convergence speed in realistic decentralised learning settings.

Comparing across datasets, the relative importance of structural and temporal heterogeneity as drivers of the slowdown reflects the nature of each deployment scenario. In the high-school network, temporal heterogeneities play a comparatively larger role than structural ones: the strong diurnal and weekly rhythms of face-to-face interaction create pronounced activity window structure that structural randomisation alone cannot recover. In the cab and WiFi networks, by contrast, structural heterogeneity is the primary limiting factor, as activity in these systems is distributed more continuously across time, making the topology of the contact graph the dominant constraint on diffusion speed.

Comparing these results with those of Section~\ref{sec:temporal}, a notable difference emerges: while in generative networks we observed a significant slowdown as a result of heterogeneous inter-event time distributions (both for the self-exciting process and renewal power-law inter-event times), their effects in slowing down real-world networks were more limited. This is visible in panels b, e, h of Fig.~\ref{fig:real-world}, where the Inter-event and Activity-constrained shufflings, which specifically target inter-event time correlations and distributions, do not produce a major acceleration compared to the original, indicating that fine-scale inter-event structure plays a negligible role in the temporal slowdown of these empirical networks. This can be traced to two structural features of the datasets and their pre-processing. First, all three datasets impose a minimum contact time cutoff, for example an event is recorded at most once per minute that two cabs are within a geographical threshold, providing a lower bound on inter-event times that dampens the most extreme effects of burstiness. This constraint also reflects a realistic feature of decentralised learning, where useful communications with the same neighbour are bounded by the frequency of local training steps. Second, the limited time window of the datasets, combined with the connectivity requirement imposed during pre-processing, effectively caps the upper tail of the inter-event time distribution. Together, these constraints push the temporal dynamics of each dataset toward a regime reminiscent of multi-scale activity patterns common in real-world systems \cite{malmgren2008poissonian, karsai2012universal}: at short time scales interactions follow a relatively regular cadence, while at longer scales the pattern remains bursty. Burstiness therefore manifests primarily as heterogeneity in link activity windows, where pairs of nodes communicate regularly within a window and rarely outside it, rather than as the heavy-tailed waiting times that most severely impair diffusion in the synthetic setting.
%
% The former constraint, in fact, provides a more realistic image of the decentralised learning setup, where useful communications with the same neighbour are bound by the frequency of local communication step. The lower-bound limit can also be viewed similar to the phenomena, common in real-world systems, where activity at different time-scales follows different temporal patterns \cite{malmgren2008poissonian, karsai2012universal}. In this case, activity follows a more regular interval at a smaller time-scale, while on a larger scale it still shows a bursty pattern. This results in the burstiness primarily manifesting itself in very limited activity windows for each pair of nodes, where the two nodes communicate consistently inside each window and almost not at all between two windows.

\section{Conclusion}\label{sec:conclusion}

Decentralised federated learning over realistic communication networks is inherently shaped by the structural and temporal heterogeneities of the underlying contact graph, which arise from the interplay of geographical constraints, communication technology, and the inhomogeneities of human mobility and interaction patterns. These heterogeneities have long been studied in the context of epidemic spreading, social network dynamics, and transport system design \cite{colizza2006role, pastor2001epidemic, wasserman1994social, granovetter1973strength, barthelemy2011spatial, badie2018error}, yet their role in decentralised federated learning has remained largely unexamined.

In this paper we have established a precise connection between decentralised federated learning and lazy random-walk diffusion on temporal networks, valid both in the early synchronisation phase and in the stationary regime where local training perturbations propagate through the network. Building on this connection, we have shown that structural and temporal heterogeneities universally and often dramatically slow down the diffusion process, with their effects compounding when multiple heterogeneities co-occur as in real-world networks. Crucially, the standard experimental setup used in decentralised federated learning research, based on homogeneous random graphs with regular communication intervals, corresponds to the fast-mixing end of the spectrum and can overestimate convergence speed by one to two orders of magnitude compared to empirically grounded scenarios. This has direct practical implications for the design and evaluation of decentralised learning protocols, suggesting that realistic benchmarks should account not only for irregular network topologies but also for the bursty, window-structured nature of real device interactions.

Several directions remain open for future work. The diffusion characterisation established here assumes linear or linearisable aggregation rules, which covers widely used algorithms such as FedAvg and its weighted variants but may not extend to arbitrary aggregation methods. A systematic study of how non-linear aggregation affects the connection to diffusion processes would broaden the applicability of our framework. More broadly, the mapping between decentralised learning dynamics and random-walk diffusion on temporal networks opens the door to leveraging the rich toolkit of network science, including mixing time bounds, spectral methods for temporal graphs, and topology-aware protocol design, to develop principled strategies for accelerating convergence in realistic heterogeneous settings.

\bibliographystyle{IEEEtran}
\bibliography{IEEEabrv,references}

\clearpage

\appendices

\section{Perturbation experiments}\label{sec:perturbation-experiment-apdx}

In this appendix we validate the diffusive relaxation approximation for the propagation of a single-node perturbation under the decentralised pairwise-averaging dynamics, using an ensemble of independent temporal realisations. We fix a static communication graph, a single realisation of $G(n, p)$ model with $n=100$ and $p=0.06$ ($\langle k \rangle = 6.46$), and assume that each undirected link $\{i, j\}$ is activated by a homogeneous Poisson process with constant (and equal) rate $\lambda_{i,j} = 1.0$ in time window $\mathcal{T} \in [0, 10)$. Across runs we keep the static graph and rates fixed but resample all activation times independently. For each temporal realisation we initialise all nodes with the same parameter vector $w^\text{(baseline)}$ except a designated source node $s$, always set to node 0, which is initialised with a second parameter vector $w^\text{(source)}$. Whenever a contact event $(\{i,j\},t)$ occurs, both nodes apply simple averaging, $w_i \leftarrow \tfrac12(w_i+w_j)$ and $w_j \leftarrow \tfrac12(w_i+w_j)$.

To quantify the spread of the source perturbation, we track for each node $i$ the source influence coefficient $\alpha_i(t)\in[0,1)$, defined as the weight of the source initialisation $w^\text{(source)}$ in the linear decomposition of $w_i(t)$ under the averaging operator. For more information about this, refer to \cref{fig:diffusion-schematic} and \cref{sec:diffusion-theoretical}. This yields an empirical trajectory $\alpha^{(r)}(t)$ for each realisation. For this first part of the experiment, we completely disregarded any effects from local training.

Under the Poisson activation model, the ensemble-average dynamics are governed by a diffusive relaxation equation $\dot{\alpha}(t)=-(1/2)L_\lambda\alpha(t)$, where $L_\lambda=D_\lambda-\Lambda$ is the rate-weighted Laplacian and $\Lambda_{ij}=\lambda_{i,j}$ on edges of the static graph and 0 otherwise. This results in a predicted influence value of
\begin{equation}
    \alpha_{\mathrm{diff}}(t)=\exp[-(1/2)L_\lambda t]\alpha(0)\,.
\end{equation}

\begin{figure*}[htb!]
    \centering
    \includegraphics[width=0.8\linewidth]{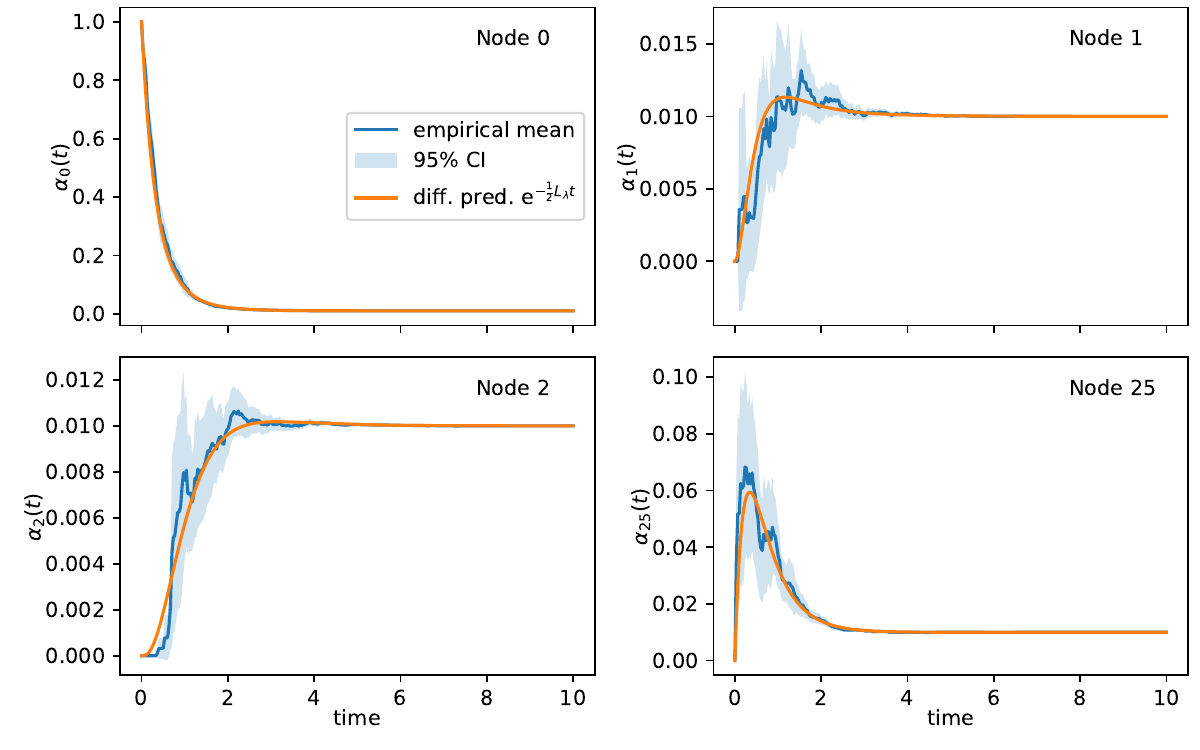}
    \caption{Influence of the source node (node 0) in the parameters of four nodes $i \in \{0, 1, 2, 25\}$ at time $t$ can be quite accurately predicted on expectation from the diffusive relaxation operator associated with the lazy random walk diffusion process. Note that node 0 is the source node, hence $\alpha_0(0) = 1$ and $\alpha_{i\neq0}(0) = 0$. Shaded area shows the 95\% confidence interval for the empirical influence across 70 realisations of the same temporal network.}
    \label{fig:influence-empirical}
\end{figure*}

In \cref{fig:influence-empirical}, for selected target nodes we report the empirical mean influence trajectory of a set of nodes $\mathbb{E}[\alpha_i(t)]$ estimated across 70 independent temporal realisations together with 95\% confidence intervals. We overlay the diffusive relaxation operator prediction $\alpha_{\mathrm{diff},i}(t)$ computed from the Laplacian of the underlying temporal graph.

Essentially, in absence of other perturbations, the evolution as a result of a single perturbation $\delta = w^\text{(source)} - w^\text{(baseline)}$ to a single source node can be understood as this: consider the line in the configuration space between the two points at $w^\text{(baseline)}$ and $w^\text{(source)}$. Right after the perturbation, the source node starts at $w^\text{(source)}$ while the rest of the nodes are at $w^\text{(baseline)}$. Barring other perturbations, at time $t$ the position of each node $i$ along that path is $w^\text{(baseline)} + \alpha_i(t)\|\delta\|$. The values $\alpha_{\mathrm{diff}}(t)$ describes the expected trajectory of the nodes, driven from the contact rates between nodes, while $\alpha_{\mathrm{emp}}(t)$ describes the actual trajectory calculated from the specific contact sequence of each realisation, in the manner described in \cref{sec:early-stage}. As can be seen in \cref{fig:influence-empirical}, across multiple runs, the empirical trajectories are described fairly accurately by the diffusive relaxation equation.

\begin{figure}[htb]
    \centering
    \includegraphics[width=0.8\linewidth]{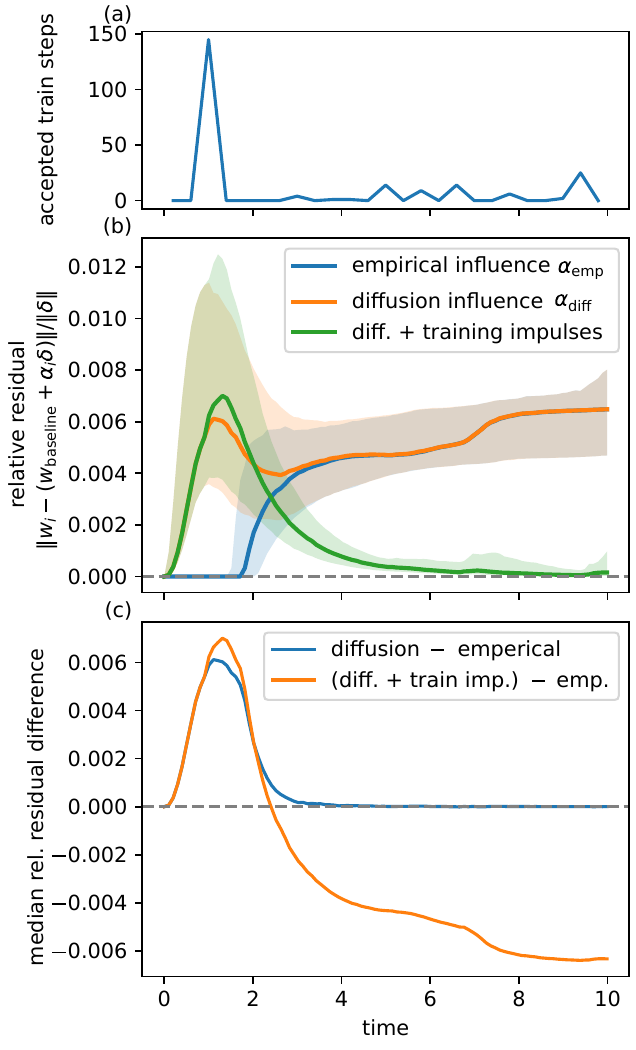}
    \caption{Residuals for parameters compared to diffusion process predictions. While $\alpha_{\mathrm{emp}}(t)$ and (on expectation) $\alpha_{\mathrm{diff}}(t)$ explain the behaviour of the system when only a single initial perturbation is analysed, additional perturbations (as a result of local training in nodes other than source) affect the system trajectory. However, accounting for each individual perturbation as an additive linear impulse on the system, diffusing based on the same diffusive relaxation kernel, removes the majority of the error in prediction (b, c). Panel (a) displays the number of accepted training steps across 70 runs. Relative residual is the magnitude of the difference between a node's predicted position based on $\alpha_{\mathrm{diff}}(t)$ or $\alpha_{\mathrm{emp}}(t)$ compared to its actual position on the configuration space, as a fraction of the magnitude of the original perturbation. Shaded area in (b) displays the 25th to 75th quantile of relative residual of nodes. Panel (c) shows that the linear addition of the original perturbation and subsequent local training impulses significantly reduces the relative residual.}
    \label{fig:residuals}
\end{figure}

To understand the effect of additional perturbations, we simulate a more realistic setup, where each node performs a local training step (a single training minibatch) at times determined through a local Poisson process with rate $\lambda_\text{train} = 0.1$. If the performed training step improves the test set loss, the new parameters are adopted for that node. For the sake of simplicity, all nodes were allocated the same set of training and test items, though each node uses a random batching of items. Importantly, unlike the previous scenario, local training steps not only move the nodes along $\delta$, but also orthogonal to it. If we take each local training step and the initial perturbation as linear impulses to the system, each defusing independently based on the same diffusive relaxation operator $-1/2L_\lambda t$, we can estimate the total effect of all impulses for each node. \cref{fig:residuals} show that this linear additive process sufficiently predicts the evolution of the system, confirming the connection to diffusion processes.

It is, however, important to note that the major difference between a diffusion process with random perturbations and the decentralised federated learning setup is based on the fact that in the federated learning setup, the local training impulses are not random or independent. To the contrary, the effect of local training for two nodes close to each other in the configuration space can be correlated with each other based on the local geometry of the loss landscape.

\subsection*{Experimental setup}

All perturbation-response experiments use the MNIST handwritten digit classification task. Inputs are $28\times 28$ greyscale images, normalised using the standard MNIST mean and standard deviation. The classifier is a small convolutional neural network consisting of two $5\times 5$ convolutional layers (16 and 32 channels, ReLU activations, each followed by $2\times 2$ max-pooling), then a fully connected layer of width 128 with ReLU and a final linear layer producing 10 logits. Optimisation uses Adam with learning rate $\eta=10^{-3}$ and weight decay $10^{-3}$. All evaluations are carried out on the standard MNIST test set.

To construct a controlled ``baseline--source'' perturbation direction, we first train two reference models offline. We sample a set of 10,000 training examples uniformly at random from the MNIST training set and train the \emph{baseline} model $w^{(\mathrm{baseline})}$ on this subset until the test loss saturates according to an early-stopping criterion (minimum number of epochs, patience window, and a minimum improvement threshold). We then continue training from the baseline weights using an \emph{augmented} dataset obtained by adding 2,000 additional i.i.d. training examples (disjoint from the first 10,000, so that the augmented set contains the original 10,000 as a subset). The resulting weights define the \emph{source} model $w^{(\mathrm{source})}$. This construction ensures that the perturbation direction $\delta = w^{(\mathrm{source})} - w^{(\mathrm{baseline})}$ corresponds to a realistic learning-induced update caused by extra data, while keeping the two models close enough in parameter space for a linear response approximation to be meaningful.

In each decentralised simulation run, we initialise all nodes $i\neq s$ with $w^{(\mathrm{baseline})}$ and initialise the source node $s$ (fixed to node 0) with $w^{(\mathrm{source})}$. The communication process follows the temporal network model described above: A single static contact network, based on an Erdős–Rényi $G(n, p)$ model with $n=100$ and $p=0.06$ describes the relationship between nodes across all runs. Each static edge ${i,j}$ of the $G(n, p)$ network is activated by an independent Poisson process with rate $\lambda_{i,j}=1$, and a contact event applies pairwise averaging to all parameters of $i$ and $j$, $w_i \leftarrow \tfrac12(w_i+w_j)$ and $w_j \leftarrow \tfrac12(w_i+w_j)$. Local training events occur at each node according to an independent Poisson process with rate $\lambda_{\mathrm{train}}=0.1$. Each local training event performs a single stochastic-gradient update (one minibatch) using Adam on the \emph{shared} 10,000-example baseline subset; i.e., all nodes have access to the same training pool, and heterogeneity in local training arises only from random minibatch selection and the current model state. After a local step, the node evaluates the updated model on the test set and adopts the update only if the test loss decreases by at least a fixed tolerance (otherwise the node reverts to its pre-update parameters). This ``accept-if-improves'' rule yields sparse, irregular training impulses that are convenient to treat as perturbations in the linear superposition analysis.

For the perturbation-response analysis, we do not store full parameter vectors for each intermediate state; instead, we record a deterministic ``fingerprint'' consisting of a fixed set of 16 model parameters. These sampled coordinates are used to compute residual norms relative to the diffusion-based predictors. The same sampling specification is used across all nodes and across all runs, ensuring that residual statistics and confidence intervals are comparable across realisations.

\section{Randomised reference models}\label{sec:randomised-reference-models}

For this work, we focus on two specific subclasses of randomised reference models, namely (1) Link shuffling and (2) timeline shuffling techniques. In broad strokes, link shuffling methods focuses on randomising structural heterogeneities, while timeline shuffling methods focus on temporal aspects of the system. Ref.~\cite{gauvin2022randomized} provides detailed formal definition and an overview of the properties of each reference model. In the following, we use the same naming scheme and canonical notation to denote each model, with the mathematical notation $\mathrm{P}[x]$ indicating a randomised reference model where value $x$ (be it a scalar, matrix or a graph) is exactly preserved in each realisation. For each randomisation, an ensemble of ten realisations were created.

\emph{Topology-constrained link shuffling} ($\mathrm{P}[\mathcal{L}, p_\mathcal{L}(\mathbf{\Theta})]$) shuffles entire timelines of events between pairs of nodes, while exactly keeping the static network projection of the temporal network fixed, meaning that whether or not two nodes have at least one event between them is preserved in the randomisation. On the other hand, \emph{Degree-constrained link shuffling} ($\mathrm{P}[\mathbf{k}, p_\mathcal{L}(\mathbf{\Theta})]$) merely preserves the exact degree sequence of the static projection, i.e., the number of neighbours each node is preserved.

In all observations, \emph{topology-constrained} and \emph{degree-constrained} link shuffling produce almost identical results but distinctly different from both the original network ($\mathrm{P}[G]$) and the unconstrained \emph{link shuffling} ($\mathrm{P}[p_\mathcal{L}(\mathbf{\Theta})]$) where entire timelines are shuffled with the only constraint that only number of static projection links are exactly preserved, with no regard for its structure. This shows that while properties such as temporal motifs might play a fairly significant role in slowing down the diffusion process on networks (since they are randomised away by all link shuffling methods), at the scale of these system we do not observe a significant effect of the system dimensionality or static network motifs, except insofar as they might affect the degree sequence or the less well connected nodes that were left out of the experiment in the pre-processing step. Such hypothetical effects would be reflected in the form of a significant difference between \emph{topology-constrained} and \emph{degree-constrained} link shuffling.

On the other hand, we note the significant effect of unconstrained \emph{link shuffling} compared to \emph{degree-constrained link shuffling}, highlighting the role of degree sequence heterogeneity. Interestingly, since \emph{link shuffling} randomises all structural heterogeneities of the systems, certain temporal patterns become easy to spot on the trajectory of inverse participation ratio over time. Specifically it is easy to observe diurnal (and for the case of high-school dataset \ref{fig:real-world}(a), weekly) patterns of activity where inactive periods (e.g., nights and weekends) are marked by plateaus.

\emph{Inter-event time shuffling} ($\mathrm{P}[\mathbf{\pi}_\mathcal{L}(\mathbf{\Delta\tau}), \mathbf{t}^1]$) independently shuffles the inter-event time sequence of each timeline, thereby randomising the possible inter-event time correlations (e.g., memory effects) and possible memory effects, as well as possible temporal motifs, while preserving the probability distribution of inter-event times as well as the time of the first event $\mathbf{t}^1$ on the timeline. By comparison, \emph{interval shuffling} ($\mathrm{P}[\mathbf{\pi}_\mathcal{L}(\mathbf{\Delta\tau})]$) also randomises the start time of the first event on each timeline, as well as the order of inter-event times. Comparing these with the original network shows that the inter-event time memory plays a fairly insignificant role in these systems, while the activity window of each timeline (the window between first and last activation) does play some role. This is further confirmed by comparing \emph{activity-constrained timeline shuffling} ($\mathrm{P}[\mathbf{\pi}_\mathcal{L}(\mathbf{\tau}), \mathbf{t}^1, \mathbf{t}^w]$), which redistributes event at random in each timeline, while preserving activity windows (time of the first and last event) of each timeline, which is in all cases only negligibly different compared to the original network.

While \emph{weight-constrained timeline shuffling} ($\mathrm{P}[\mathbf{w}]$) and the unconstrained \emph{timeline shuffling} ($\mathrm{P}[\mathcal{L}]$), similar to all timeline shuffling methods mentioned before, both preserve the underlying static network projection of the system, they randomise activation window and inter-event time distribution of the network. The major difference stems from the fact that \emph{weight-constrained timeline shuffling} additionally constrains the exact number of activations of each link between two nodes. As we see in the case of cab and Wi-Fi network, this heterogeneity of activation frequencies can potentially play a major role in the rapidity of the diffusion process.

Combining temporal and structural randomisation methods, i.e., consecutive application of \emph{link} and \emph{timeline shuffling}, shows that in all cases structural and temporal heterogeneities both independently contribute to the slowing down of the diffusion. The resulting randomised network is essentially a $G(N, m)$ Erdős–Rényi model network (with same number of nodes and links as the original) and exactly the same total number of activations as the original network, but distributed randomly at random times among the $M$ links, resulting in an exponential inter-event time distribution. In the case of high-school network, temporal heterogeneities play a comparatively larger role in slowing down the network than the structural ones, while on the other two datasets the opposite is true.

\end{document}